\documentclass[final]{cvpr}

\usepackage{epsfig}
\usepackage{amssymb}

\usepackage{times}  
\usepackage{helvet} 
\usepackage{courier}  
\usepackage[hyphens]{url}  
\usepackage{graphicx} 
\usepackage{caption} 
\usepackage{amsmath}
\usepackage{bm}
\usepackage{lipsum,multicol}
\usepackage{subcaption}
\usepackage{multirow}
\usepackage{mathrsfs}
\usepackage{soul}
\usepackage{booktabs}
\usepackage{graphicx}
\usepackage[dvipsnames]{xcolor}
\usepackage{algorithm}  
\usepackage{algorithmicx} 
\usepackage{algpseudocode}
\usepackage{wrapfig} 
\usepackage{makecell}

\usepackage[pagebackref=true,breaklinks=true,colorlinks,bookmarks=false]{hyperref}

\def\cvprPaperID{10917} 
\def\confYear{CVPR 2021}

\begin{document}

\title{Pareto Self-Supervised Training for Few-Shot Learning}

\author{Zhengyu Chen\textsuperscript{\rm 1,2,3},
	Jixie Ge\textsuperscript{\rm 1,2}, Heshen Zhan\textsuperscript{\rm 1,2}, Siteng Huang\textsuperscript{\rm 1,2},
	Donglin Wang\textsuperscript{\rm 1,2}\thanks{{Corresponding author.} } \\
	\textsuperscript{\rm 1}  Machine Intelligence Lab (MiLAB), AI Division, School of Engineering, Westlake University \\
\textsuperscript{\rm 2} Institute of Advanced Technology, Westlake Institute for Advanced Study   \\
\textsuperscript{\rm 3} College of Computer Science \& Technology, Zhejiang University \\
{\tt\small \{chenzhengyu,gejixie,zhanheshen,huangsiteng,wangdonglin\}@westlake.edu.cn}
}

\maketitle
 
\begin{abstract}
While few-shot learning (FSL) aims for rapid generalization to new concepts with little supervision, self-supervised learning (SSL) constructs supervisory signals directly computed from unlabeled data. Exploiting the complementarity of these two manners, few-shot auxiliary learning has recently drawn much attention to deal with few labeled data. Previous works benefit from sharing inductive bias between the main task (FSL) and auxiliary tasks (SSL), where the shared parameters of tasks are optimized by minimizing a linear combination of task losses. However, it is challenging to select a proper weight to balance tasks and reduce task conflict. To handle the problem as a whole, we propose a novel approach named as Pareto self-supervised training (PSST) for FSL. PSST explicitly decomposes the few-shot auxiliary problem into multiple constrained multi-objective subproblems with different trade-off preferences, 
and here a preference region in which the main task achieves the best performance is identified. Then, an effective preferred Pareto exploration is proposed to find a set of optimal solutions in such a preference region. Extensive experiments on several public benchmark datasets validate the effectiveness of our approach by achieving state-of-the-art performance.

\end{abstract}


\section{Introduction}

Although deep learning has achieved great success in a variety of fields, limitations still exist in many practical applications where labeled samples are intrinsically rare or expensive. 
Different from humans who can easily learn to accomplish new tasks with a few examples, it is difficult for machines to rapidly generalize to new concepts with very little supervision, which draws considerable attention to the challenging \textit{few-shot learning} (FSL) setting. As training large models with few labeled samples leads to overfitting or even non-convergence, conventional deep neural networks fail to address such a problem. 

Recently, \textit{self-supervised learning} (SSL) attracts many researchers for its soaring performance without involving manual labels. By defining pretext tasks to exploit the structural information of data itself, supervisory signals can be easily developed to learn useful general-purpose representations~\cite{bojanowski2017unsupervised,he2020momentum,pathak2016context}. As self-supervised learning can improve the generalization of the network under the limitation of labeled data, some recent \textit{few-shot auxiliary learning} (FSAL) works~\cite{GidarisBKPC19,SuMH20} take few-shot learning as learning main task with self-supervised auxiliary tasks. To encourage that the few-shot task benefits from auxiliary tasks, some parameters are shared across tasks to inductive knowledge transfer. However, as the objectives of distinct tasks are different and the relationship between objectives is complicated and unknown, optimizing each task not only promotes each other but also naturally conflicts. A typical solution to suppress such conflict is to optimize the shared parameters by minimizing a weighted sum of the empirical risk for each task, where each weight of the empirical risk can be viewed as the trade-off. 
In previous works~\cite{GidarisBKPC19,SuMH20}, these trade-offs are usually set by experience in practical situations.
It is difficult to find optimal trade-offs. 
Moreover,  
these works~\cite{GidarisBKPC19,SuMH20} attempt to find one single solution for all objectives, which is likely to sacrifice the performance of the main task and be inconsistent with the goal of few-shot auxiliary learning.

\begin{figure}
	
	\begin{subfigure}{0.46\linewidth}
		\includegraphics[width=\linewidth]{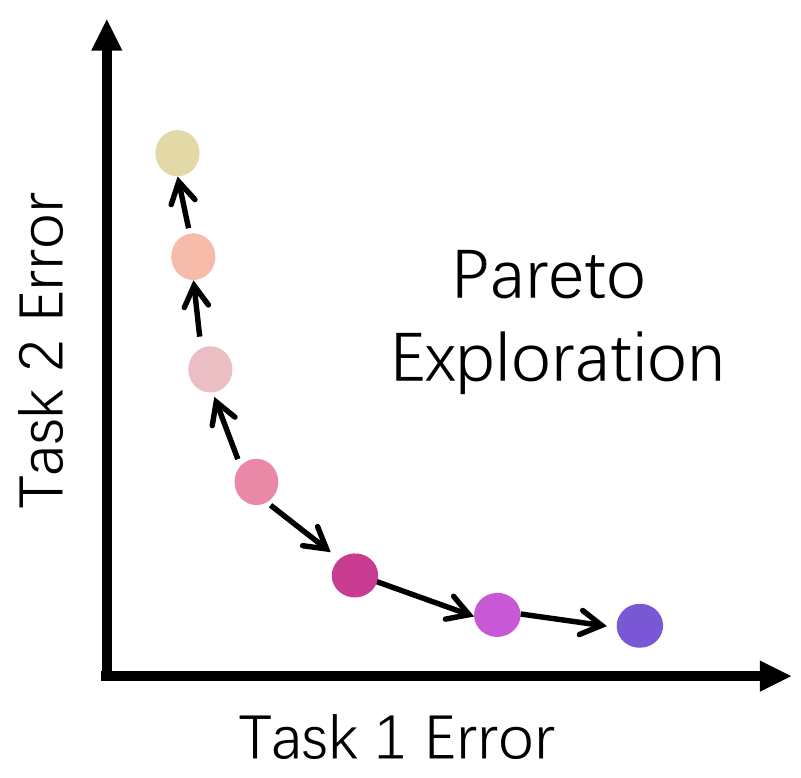}\label{ablation}
 
		\caption{}
	\end{subfigure}
	\begin{subfigure}{0.46\linewidth}
		\includegraphics[width=\linewidth]{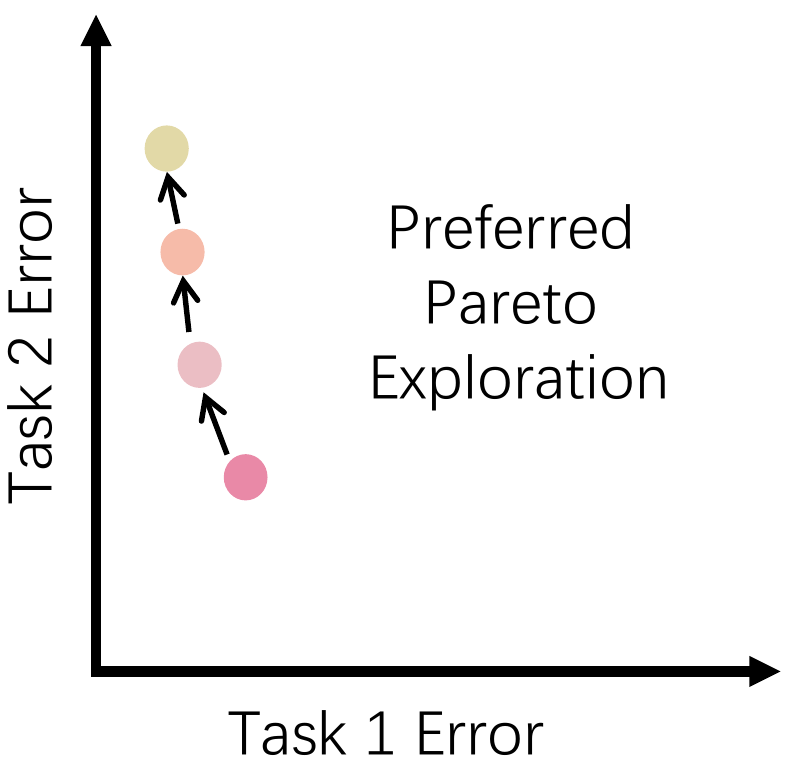}\label{entropy}
 
		\caption{}
	\end{subfigure}
	\caption{Illustrative examples of Pareto exploration in (a) previous works and (b) our PSST.}\label{figtsn}
\end{figure}

According to the above discussion, few-shot auxiliary learning with conflicting objectives requires better modeling of the trade-off between tasks, which is beyond what a linear combination achieves. To overcome the issue, we propose a novel approach named \textbf{Pareto self-supervised training (PSST)} for few-shot learning. PSST explicitly casts few-shot auxiliary learning as a multi-objective optimization problem, with the overall objective of finding a Pareto optimal solution of network parameters \cite{LinZ0ZK19,ma2020continuous}.  
However, different from previous works that explore in the global space {\cite{ma2020continuous}}, PSST uses an effective preferred Pareto exploration for FSL. Specifically, PSST decomposes the few-shot auxiliary problem into several constrained multi-objective subproblems with different trade-off preferences, and then identifies the preference region where the main task achieves the best performance. As illustrated in Fig.~\ref{figtsn}, the desired space of exploration is thus restricted by only exploring in the preference region where the given points achieve better performance in task 1 rather than task 2. Experiments demonstrate that this improvement can suppress the accumulation of residual error, which contributes to efficiently finding a more accurate Pareto solution. To summarize, our main contributions are as follows:

\begin{itemize}
	\item We point out that existing few-shot auxiliary learning methods face with a linear combination of conflicting objectives, and propose a multi-objective optimization solution to address the issue.
	\item We propose a novel Pareto self-supervised training (PSST) approach for few-shot auxiliary learning. To achieve better performance for the main task and meanwhile improve the efficiency and accuracy of the exploration, PSST pioneers a preferred Pareto exploration that explores in the identified preference region.
	\item We conduct extensive experiments to demonstrate that our PSST can better model the trade-off between tasks, which leads to state-of-the-art performance on several benchmark datasets.
	
\end{itemize}

\section{Related Work}

\subsection{Few-Shot Learning}

Few-shot learning aims to generalize well to the novel classes where only a few labeled samples are available. Recently, \textit{meta-learning} has been considered as the main solution to the few-shot problem due to the significant progress. Regarded as ``learning to learn'', meta-learning aims to improve its future learning performance with the experience of multiple learning episodes. Current meta-learning approaches for the few-shot problem can be roughly divided into three groups: optimization-based, model-based, and metric-based. \textit{Optimization-based} approaches~\cite{chenmu21,ChenWang21,FinnAL17} learn a meta-learner to adjust the optimization algorithm, usually by providing better initialization or search steps for parameters. \textit{Model-based}~\cite{BertinettoHTV19,GidarisK18,GuoC20,MunkhdalaiY17} approaches depend on well-designed models, whose parameters are obtained with its internal architecture or a meta-learner for fast learning. \textit{Metric-based} approaches~\cite{oreshkin2018tadam,SnellSZ17,Sung:RelationNet} learn a generalizable embedding model to transform all samples into a common metric space, where specific distance measures can be employed with the nearest neighbor classifiers.

\subsection{Self-Supervised Learning}

As collecting enough human-annotated labels for large-scale unlabeled data is difficult and expensive, self-supervised learning methods aim to learn representations from unlabeled data. In computer vision applications, various pretext tasks have been utilized for pre-training the network including relative patch location~\cite{doersch2015unsupervised}, rotation prediction~\cite{gidaris2018unsupervised}, image inpainting~\cite{pathak2016context}, and clustering~\cite{caron2019unsupervised}. Another line of works learns representations by contrasting positive pairs against negative pairs constructed on augmented samples~\cite{bojanowski2017unsupervised,dosovitskiy2014discriminative,he2020momentum,xiao2019dynamic,xiao2018neural,yan2020pointasnl}.  

Regarded as auxiliary tasks, self-supervised learning can also be used to improve other tasks. For example, \cite{zhai2019s4l} shows that self-supervision can contribute to the recognition in a semi-supervised setting, and \cite{carlucci2019domain} uses self-supervision to improve domain generalization. Recently, few-shot auxiliary learning that exploits the complementarity of both few-shot learning and self-supervised learning has drawn much attention. Showing that the auxiliary loss without labels can extract discriminative features for few-shot learning, \cite{GidarisBKPC19} considers rotation prediction and relative patch location as self-supervised tasks, and \cite{SuMH20} uses image jigsaw puzzle. In our work, instead of being limited to specific self-supervised auxiliary tasks, we propose a general and effective preferred Pareto exploration that applies to arbitrary auxiliary tasks.

\subsection{Multi-Objective Optimization}

Multi-objective optimization~\cite{zuluaga2013active} aims at finding a set of Pareto solutions with different trade-offs rather than one single solution. It has been used in many machine learning applications such as reinforcement learning~\cite{van2014multi}, Bayesian optimization~\cite{hernandez2016predictive} and neural architecture search~\cite{elsken2018efficient}. As the gradient information is usually not available in these applications, population-based and gradient-free multi-objective evolutionary algorithms~\cite{daglib,Zitzler99} are popular methods to find a set of well-distributed Pareto solutions in a single run. However, it can not be used for solving large-scale and gradient-based multi-task learning (MTL) problems. Multi-objective gradient descent~\cite{fliege2000steepest} is an efficient approach for multi-objective optimization when gradient information is available. \cite{sener2018multi} proposes a novel method for solving MTL by treating it as multi-objective optimization. \cite{LinZ0ZK19} presents an MGDA-based method to generate a discrete set of solutions evenly distributed on the Pareto front. And \cite{ma2020continuous} proposes to replace discrete solutions with continuous solution families, allowing for a much denser set of solutions and continuous analysis on them. However, these methods still focus on how to find a Pareto solution where the performance of each task is equally important. How to effectively find a Pareto solution that is concerned only with the performance of the main task still remains a challenge for auxiliary learning.

\begin{figure*}[t]
	\centering 
	\includegraphics[width=17cm]{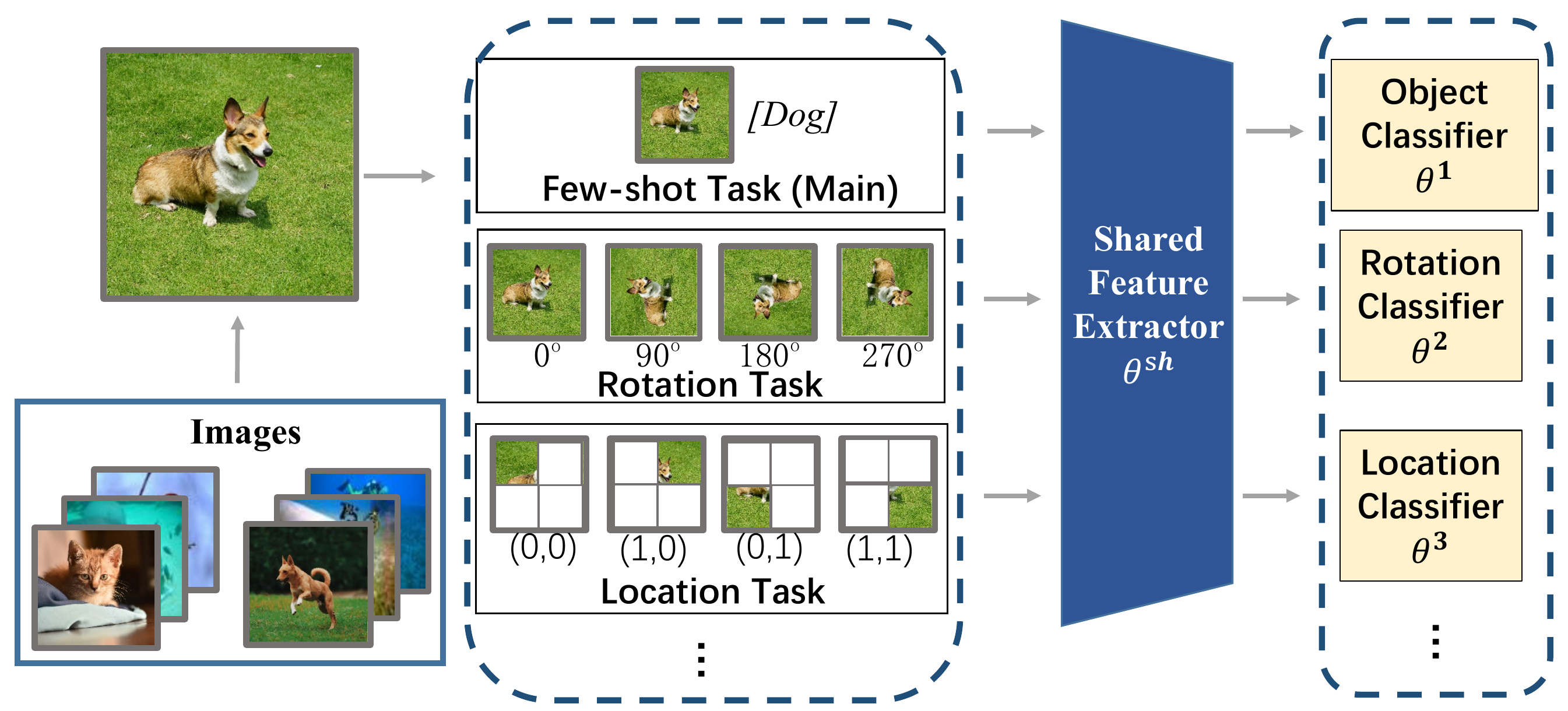} 
	\caption{The framework used in our Pareto self-supervised training.}
	\label{fig:frame}
\end{figure*}

\section{Pareto Self-Supervised Training}

In this section, we firstly introduce a general framework to state our problem to be solved. Then we formulate the problem as a multi-objective optimization for FSAL. After that, we elaborate how to solve
the optimization problem.

\subsection{Framework} 
The general framework in Fig. \ref{fig:frame} is used in our PSST for problem clarification, which combines self-supervised task and supervised class recognition task in a few-shot setting. We train the feature extractor $\theta^{sh}$ with both labeled (top branch) and unlabeled (bottom branches) data in a multi-task setting. We use the labeled data to train the object classifier $\theta^1$ with few-shot classification loss. For the self-supervised task, we sample images from the unlabeled dataset. For example, we generate four rotations for each input image in rotation task, process them with $\theta^{sh}$ and train the rotation classifier $\theta^2$ with the self-supervised loss. The pipeline for other self-supervised tasks is analog to this one.

\subsection{ Multi-Objective FSAL}

Our goal is to improve few-shot learning performance via self-supervised training.
Different from previous works using a weighted sum of loss functions for few-shot auxiliary learning with self-supervised learning \cite{DvornikMS19,GidarisBKPC19}, 
we cast few-shot auxiliary learning as multi-objective optimization rather than a weighted sum of loss function.

We consider the few-shot auxiliary learning over an input space $\mathcal{X}$ and a collection of task spaces
$\left\{\mathcal{Y}^{m}\right\}_{m \in[M]}$, such that a large dataset of independent and identically distributed (i.i.d.) data points $\left\{\mathbf{x}_{i}, y_{i}^{1}, \ldots, y_{i}^{M}\right\}_{i \in[N]}$ is given,
 where $M$ is the number of tasks, $N$ is the number of data points, and $y_{i}^{m}$ is the label of the $m^{th}$ task for the $i^{th}$ data point.
As shown in Fig. \ref{fig:frame}, we further consider a parametric hypothesis class per task as $f^{m}\left(\mathrm{x} ; {\theta}^{s h}, {\theta}^{m}\right): \mathcal{X} \rightarrow \mathcal{Y}^{m}$, such that some parameters $\left(\theta^{s h}\right)$ are shared between tasks and others $\left(\theta^{m}\right)$ are task-specific. Define the task-specific loss function by $\mathcal{L}_{m}(\cdot, \cdot): \mathcal{Y}^{m} \times \mathcal{Y}^{m} \rightarrow \mathbb{R}^{+}$. 

Previous works \cite{DvornikMS19,GidarisBKPC19} are to optimize a proxy objective that minimizes a weighted linear combination of per-task losses:
\begin{equation}
\min _{\theta} {L}(\theta)=\sum_{m=1}^{M} {\omega}_{m} {L}_{m}(\theta^{s h},{\theta}^{m}),
\end{equation}
where $\omega_m$ is the weight for the $m$-th task, and ${{L}}_{m}\left({\theta}^{s h}, {\theta}^{m}\right)$ is the empirical loss of the task $m$, defined as ${{L}}_{m}\left({\theta}^{s h}, {\theta}^{m}\right) \triangleq \frac{1}{N} \sum_{i=1}^N {L}_m\left(f^{m}\left(\mathbf{x}_{i} ; {\theta}^{s h}, {\theta}^{m}\right), y_{i}^{m}\right)$. 
This approach is simple and straightforward, however, it has some drawbacks which have been pointed out by many recent works \cite{LinZ0ZK19,ma2020continuous}.  

In a typical few-shot auxiliary learning application, the weight $\omega_m$ is needed to be assigned manually before optimization, and the overall performance is highly dependent on the assigned weights. Choosing a proper weight vector could be very difficult even for an expert.
To accurately model the trade-off between tasks, which is beyond what a linear combination can achieve, multi-objective optimization is widely used in recent works \cite{LinZ0ZK19,ma2020continuous}.
 
Previous works focus on weighted summation due to its intuitively appealing \cite{DvornikMS19,GidarisBKPC19,ChenDTTD2021},
however, these works typically require either an expensive grid search over various scalings or the use of some heuristics.
Recent work formulates the weighted summation as multi-objective optimization, which optimizes a collection of possibly conflicting objectives \cite{LinZ0ZK19}. 
By adopting this, we reformulate our few-shot auxiliary learning  as multi-objective optimization problem with  
a vector-valued loss $\mathcal{L}$:
\begin{equation}
\min _{{\theta}} \mathcal{L}\left({\theta} \right) = \min _{{\theta}} \left({\mathcal{L}}_{1}\left({\theta}^{s h}, {\theta}^{1}\right), \ldots, {\mathcal{L}}_{M}\left({\theta}^{s h}, {\theta}^{M}\right)\right)^{\top},  
\end{equation}
where
${\theta} = \left\{ {\theta}^{s h},{\theta}^{1}, \ldots, {\theta}^{M} \right\}$. The goal of multi-objective optimization is to reach Pareto optimality.

\textbf{Definition 1} (Pareto dominance): A solution $\theta$ dominates a solution $\bar{\theta}$ if ${\mathcal{L}}_{m}\left({\theta}^{s h}, {\theta}^{m}\right) \leq {\mathcal{L}}_{m}\left(\overline{{\theta}}^{s h}, \overline{{\theta}}^{m}\right)$ for all tasks $m$ and $\mathcal{L}\left({\theta}^{s h}, {\theta}^{1}, \ldots, {\theta}^{M}\right) \neq \mathcal{L}\left(\overline{{\theta}}^{s h}, \overline{{\theta}}^{1}, \ldots, \overline{{\theta}}^{M}\right)$.

\textbf{Definition 2} (Pareto optimality): A solution $\theta^{*}$ is called Pareto optimal if there exists no solution $\theta$ that dominates $\theta^{*}$.
The set of Pareto optimal solutions is called the Pareto front $\left(\mathcal{P}=\{\mathcal{L}(\theta^{*})\}_{{\theta^{*}} \in \mathcal{P}_{\theta^{*}}}\right)$. 

\begin{figure}
	\centering
	\begin{subfigure}{0.46\linewidth}
		\includegraphics[width=\linewidth]{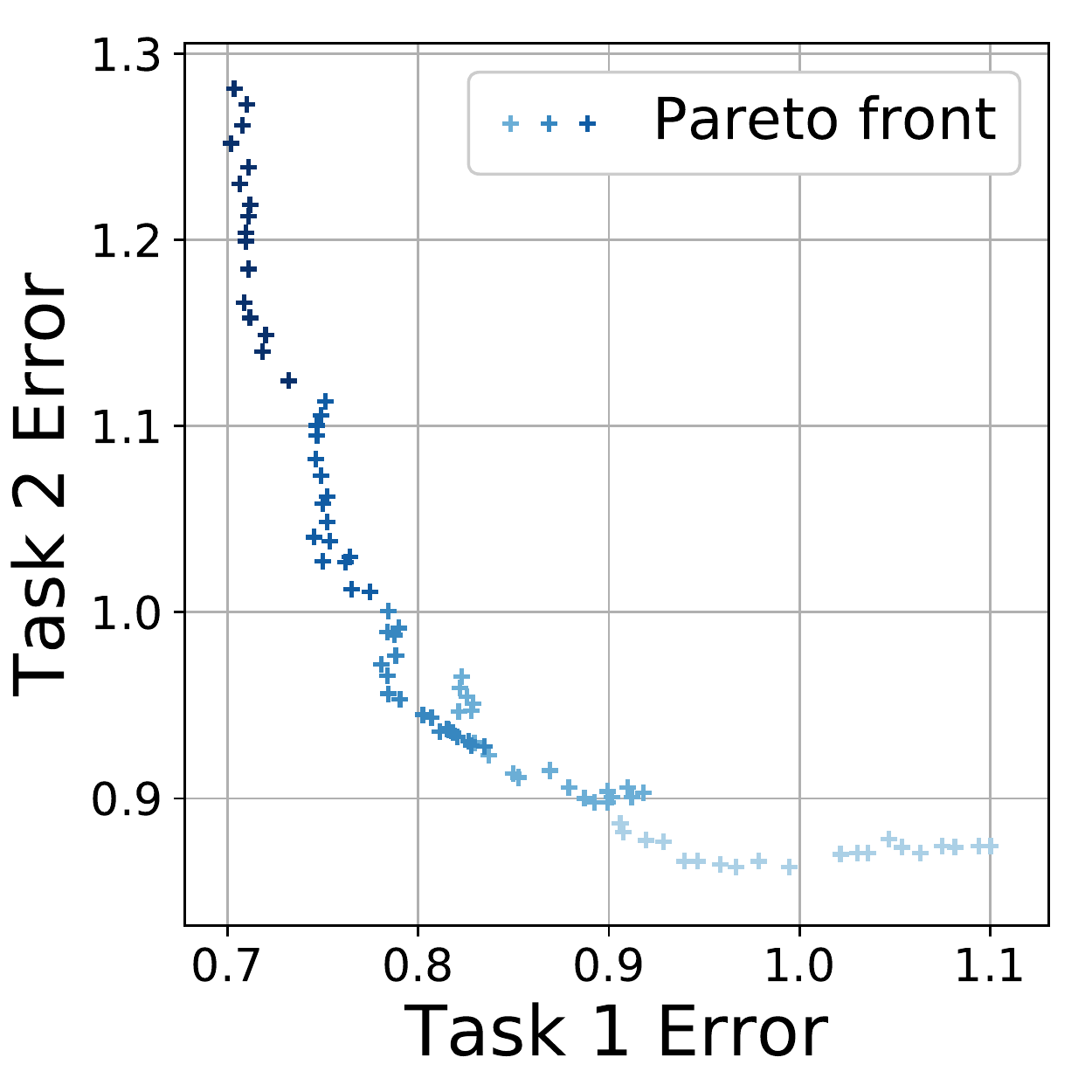}\label{common}
		\caption{Pareto front }
	\end{subfigure}
	\begin{subfigure}{0.46\linewidth}
 
		\includegraphics[width=\linewidth]{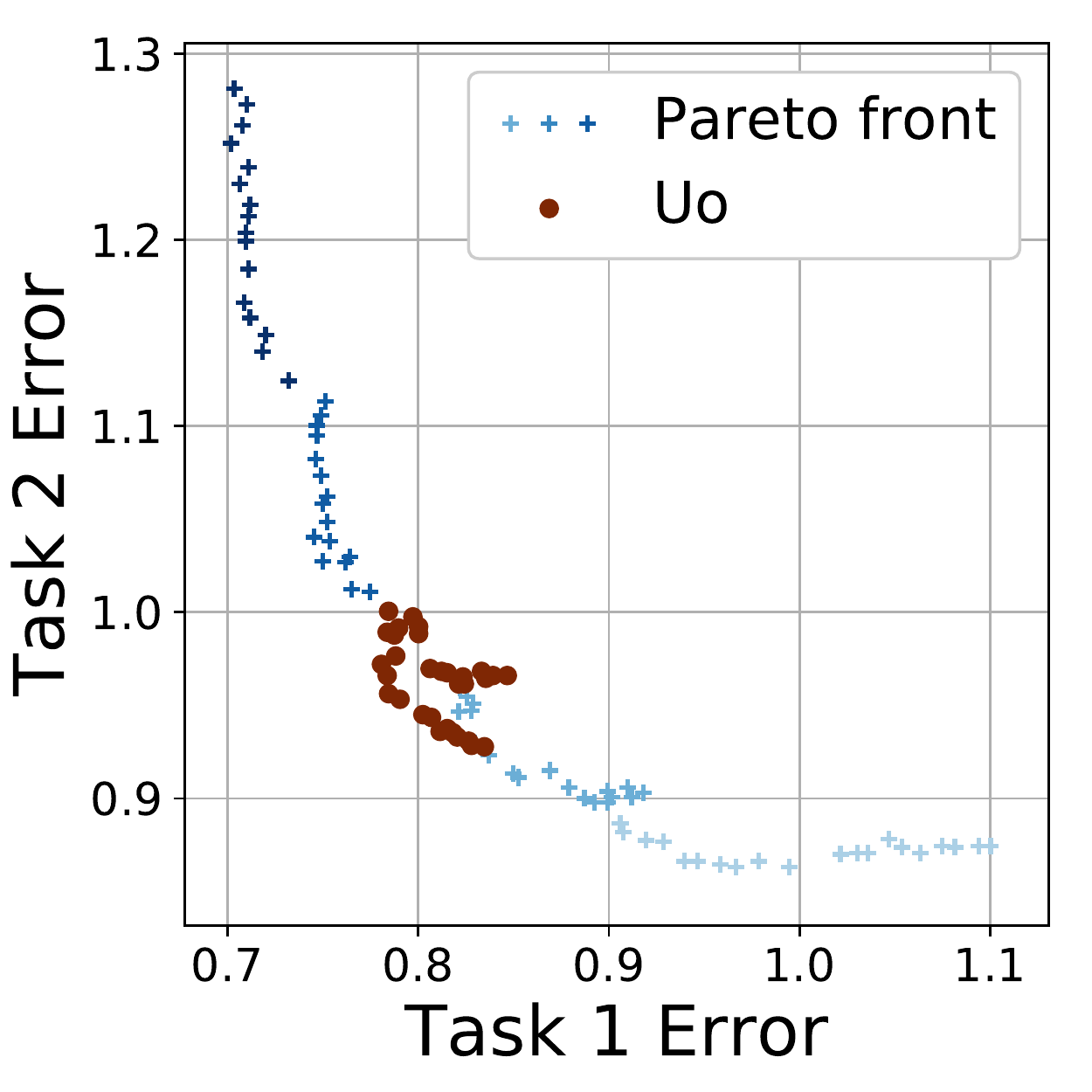}\label{private}
		\caption{$U_0$ points in Pareto front.}
	\end{subfigure}
	\caption{The convergence behaviors of PSST. Each $U_0$ point is the solution of PSST from different initial parameters of neural network.
		The proposed PSST successfully find $U_0$ in the balance space that maximizes the overall performance of two tasks. }\label{figconvergence}
\end{figure}

\subsection{Optimization of Multi-Objective FSAL}
\subsubsection{Gradient-based multi-objective optimization}

Recent works have proposed several gradient-based methods \cite{FliegeV16,LinZ0ZK19} for multi-objective optimization problems. 
The update rule of the simple gradient-based method \cite{fliege2000steepest} is $\theta_{t+1}=\theta_{t}+\eta d_{t}$, where $\eta$ is the step size and the search direction $d_{t}$ is obtained as follows:
\begin{equation}\label{eqP}
	\begin{array}{c} \left(d_{t}, \alpha_{t}\right)=\arg \min _{d \in R^{M}, \alpha \in R} \alpha+\frac{1}{2}\|d\|^{2}, \\ \text { s.t. } \quad \nabla \mathcal{L}_{m}\left(\theta_{t}\right)^{T} d \leq \alpha, m=1, \ldots, M. \end{array}
\end{equation}

The solutions of the above problem will satisfy:

\textbf{Lemma 1} \cite{fliege2000steepest}: Let $\left(d^{k}, \alpha^{k}\right)$ be the solution of problem in Eq. \ref{eqP}.
If $\theta_{t}$ is Pareto optimal, then $d_{t}=0 \in \mathbb{R}^{n}$ and $\alpha_{t}=0$. If $\theta_{t}$ is not Pareto optimal, then
\begin{equation}
	\begin{array}{c}\alpha_{t} \leq-(1 / 2)\left\|d_{t}\right\|^{2}<0, \\ \nabla \mathcal{L}_{m}\left(\theta_{t}\right)^{T} d_{t} \leq \alpha_{t}, m=1, \ldots, M, \end{array}
\end{equation}
where $\theta$ is called Pareto optimal if no other solution has better value in all objective functions.  
 
As shown in Fig. \ref{figconvergence},
 using gradient-based method in Eq. \ref{eqP} to solve multi-objective optimization problems, the interaction between gradients within their shared parameters is balanced, which leads to balanced performances for different tasks. Thus, the solutions where given multiple different initial parameters achieve balanced performances for different tasks, and these solutions are in the space where different tasks are balanced.

\begin{figure}
	\centering
	\begin{subfigure}{0.46\linewidth}
		\includegraphics[width=\linewidth]{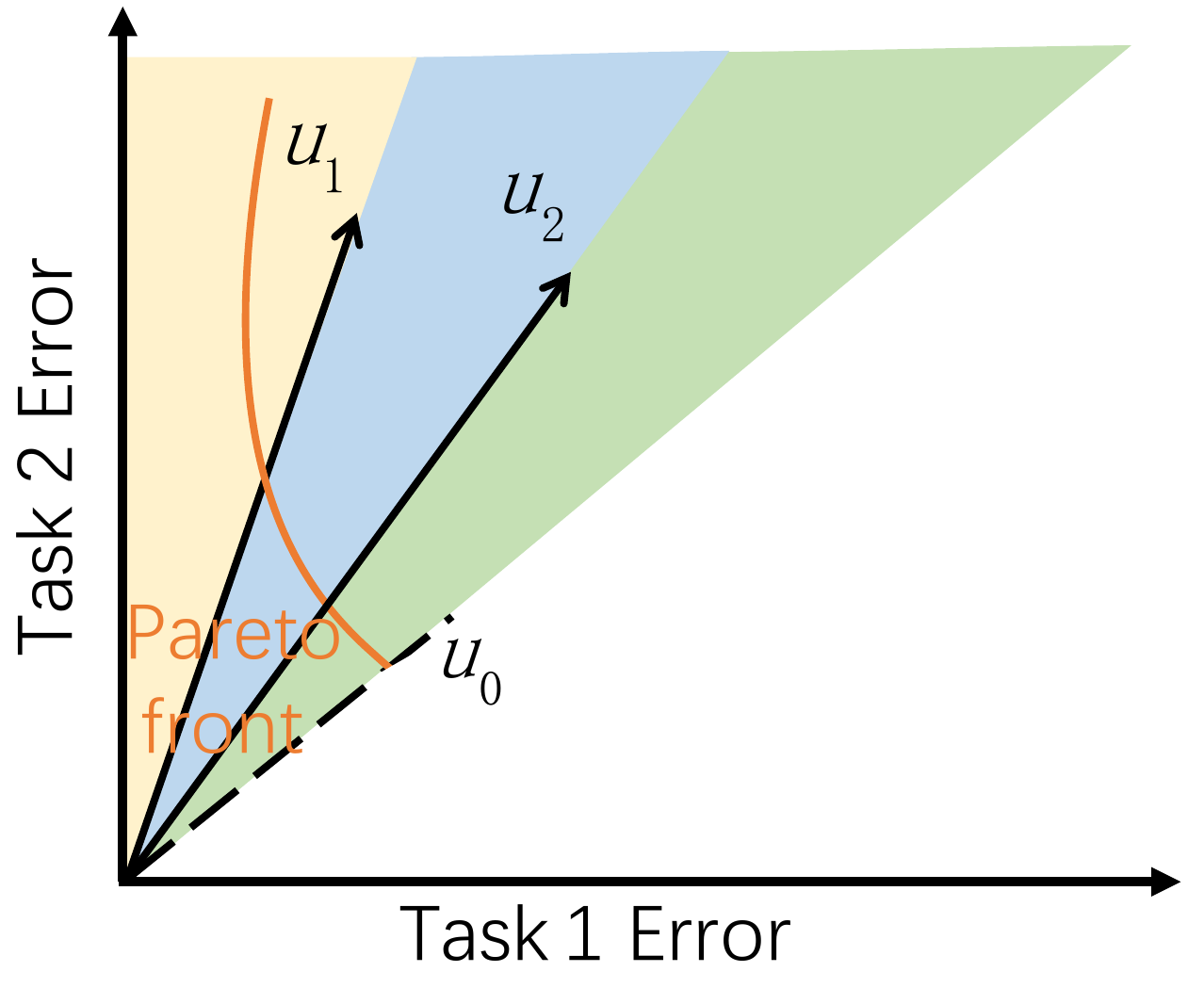}\label{common}
		\caption{}
	\end{subfigure}
	\begin{subfigure}{0.46\linewidth}
		\includegraphics[width=\linewidth]{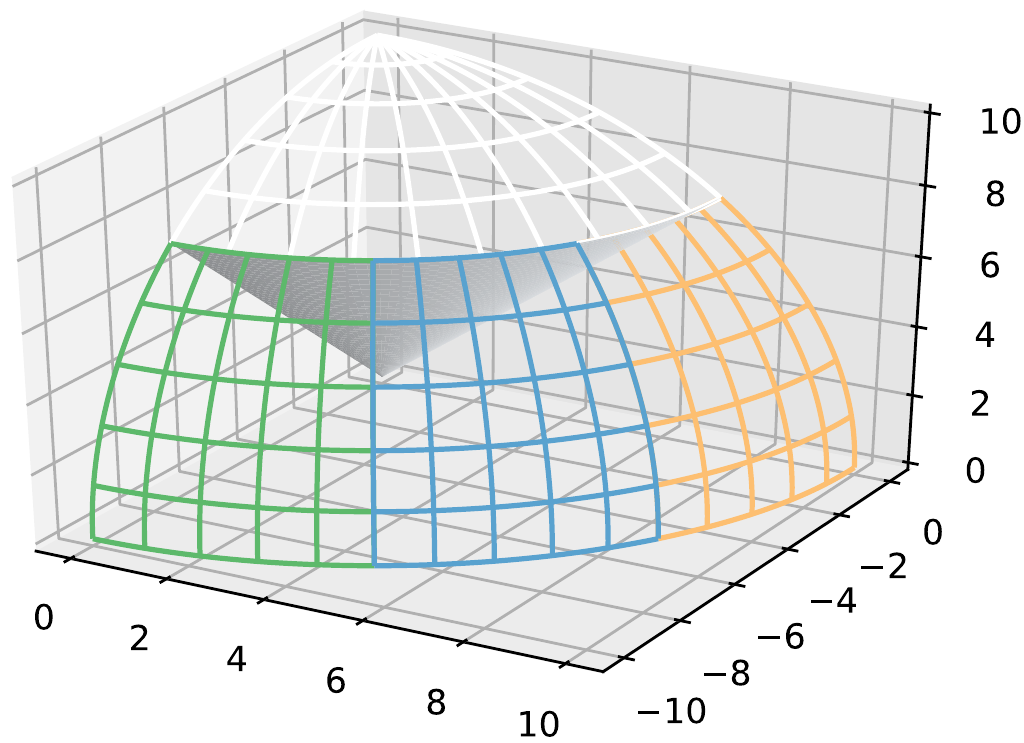}\label{private}
		\caption{ }
	\end{subfigure}
	\caption{Illustrative examples of preference region in (a) 2 tasks (b) 3 tasks, where the white space will not be explored.}\label{figpre}
\end{figure}

\subsubsection{Optimization in preference region}\label{secRegion}

Decomposition-based multi-objective evolutionary algorithm \cite{TrivediSSG17,ZhangL07} is one of the most popular gradient-free multi-objective optimization methods, which decomposes a multi-objective optimization problem into $K$ subproblems via preference vectors $\left\{\boldsymbol{u}_{1}, \boldsymbol{u}_{2}, \ldots, \boldsymbol{u}_{K}\right\}$ in $R_{+}^{M}$ and solves them simultaneously. The preference vectors in these methods are handcrafted, and all subproblems are solved.
However, in few-shot auxiliary learning, we focus on the main task, the challenge lies in how to identify the preference region where the main task achieves the best performance. 

Because of the distinct optimization of different tasks, it is difficult to directly determine the space where the main task achieves optimal performance.
Therefore, the goal is to exclude the parts that are not interested, where the auxiliary tasks hamper the performance of the main task.
We try to find the space where distinct tasks are balanced.
If the performance of the main task is not better than the balanced performance of the main task, we will remove this part in the space.
We identify and empirically demonstrate the solution of gradient-based optimization method for multi-objective optimization lying in the set where the overall performance of different tasks are balanced.  
An illustrative example is shown in Fig. \ref{figpre}.

Assuming the best overall performence is achieved when all tasks are balanced, we state that if the performance of the main task is no better than the balanced performance of the main task, the auxiliary tasks hamper the main task. Let $\rho(\theta)$ be the proportion of performance in the main task over the performance of all auxiliary tasks in $\theta$, i.e., $\rho(\theta) = \frac{\mathcal{L}_1(\theta) }{\sum_{m=2}^{M} \mathcal{L}_m(\theta) } $, we formalize the result as follows.
\textbf{Lemma 2}: If $\theta_{\pi0}^{*}$ achieves the best overall performance, i.e. $min_{\theta} \sum_{m=1}^{M} \mathcal{L}_m(\theta) = \sum_{m=1}^{M} \mathcal{L}_m(\theta_{\pi0}^{*})$, and a better performance of the main task can be achieved in $\theta'$, i.e., $ \mathcal{L}_1(\theta') < \mathcal{L}_1(\theta_{\pi0}^{*})$,  then $\rho(\theta') < \rho(\theta_{\pi0}^{*}) $.
 
Noting that $\rho(\theta) = \rho(\theta_{\pi0}^{*}) $ is a hyperplane dividing the objective space into two parts, we remove the part satisfying $\rho(\theta) > \rho(\theta_{\pi0}^{*}) $, and further divide the remaining part satisfying $\rho(\theta) \leq \rho(\theta_{\pi0}^{*}) $ into different regions with preference vectors to avoid residual error accumulation, which is detailed in Section \ref{paretoEx}. An illustrate example is shown in Fig. \ref{figpre}, the white space stands for $\rho(\theta) > \rho(\theta_{\pi0}^{*})$.

\textbf{Subproblem Decomposition.}
With the solution $\theta_{\pi0}^*$, we have 
$\mathcal{L}(\theta_{\pi0}^*)=[\mathcal{L}_{1}(\theta_{\pi0}^*), \mathcal{L}_{2}(\theta_{\pi0}^*), \cdots, \mathcal{L}_{M}(\theta_{\pi0}^*)]^{\mathrm{T}}$.
In a 2-D space, $\rho (\theta)=\rho (\theta^*_{\pi 0})$ is a line, its direction vector is well defined. In a M-dimentional (M-D) space ($M\geq3$), $\rho (\theta)=\rho (\theta^*_{\pi 0})$ is a hyperplane. This hyperplane will intersects the coordinate planes and form some lines. We define the direction vectors of $\rho (\theta)=\rho (\theta^*_{\pi 0})$ in M-D space as the direction vectors of the intersections formed by the hyperplane and coordinate planes. Coordinate plane is 2-D space, which corresponding two tasks with the name ``two tasks scenario" comes from this place.
Let unit vector $u_0$ be one of the direction vector of $\rho(\theta) = \rho(\theta_{\pi0}^*)$ for two-task scenario, we have  $u_0 = \left(\cos \pi_{0}, \sin \pi_{0} \right)$, where $\cos \pi_0 = \frac{e_{1} \mathcal{L}(\theta_{\pi0}^*)}{\left \| T_2 \mathcal{L}(\theta_{\pi0}^*)\right \|_2}$ and $T_m = I - e^{mm}$. 
Here, $e^{mm}$ is a single-entry matrix, i.e. the $m$th element in the $m$th column being one and the rest elements being zero; $e_{1} = (1 , 0)^T$; $I$ is an identity matrix. Given the $u_0$, we further decompose the auxiliary problem into $K$ subproblems with a set of unit preference vectors  $\left\{\boldsymbol{u}_{1}, \boldsymbol{u}_{2}, \ldots, \boldsymbol{u}_{K}\right\}$, for the preference vector $u_i$:
\begin{equation}\label{ui}
\begin{array}{c}
u_i = \left(\cos \pi_{i}, \sin \pi_{i} \right), \\
s.t. \quad \pi_{i} = \frac{i}{K} \left(\frac{\pi}{2} - \pi_0 \right) + \pi_0, i = 1, \ldots, K, \\
\end{array}
\end{equation}

Suppose all objectives in the multi-objective optimization are non-negative, the multi-objective subproblem corresponding to the preference vector $u_i$ and $u_{i+1}$ is:
\begin{equation}\label{subproblem} \begin{array}{c} \min _{\theta} \mathcal{L}(\theta)=\left(\mathcal{L}_{1}(\theta), \mathcal{L}_{2}(\theta), \cdots, \mathcal{L}_{M}(\theta)\right)^{\mathrm{T}}, \\  \text { s.t. }  u_i {e_1}^T  \leq \frac{e_1 \mathcal{L}(\theta)}{\left \| T_2 \mathcal{L}(\theta) \right \|_2  }  \leq u_{i+1} {e_1}^T. 
\end{array} \end{equation} 

The subproblem Eq.\ref{subproblem} can be further reformulated as:
\begin{equation} \begin{array}{c} \min _{\theta} \mathcal{L}(\theta)=\left(\mathcal{L}_{1}(\theta), \mathcal{L}_{2}(\theta), \cdots, \mathcal{L}_{M}(\theta)\right)^{\mathrm{T}}, \\  \text { s.t. }  \mathcal{Q}_i(\theta) = \frac{e_1 \mathcal{L}(\theta)}{\left \| T_2 \mathcal{L}(\theta) \right \|_2  } - u_{i+1} {e_1}^T \leq 0, \\
\mathcal{R}_i(\theta) = u_i {e_1}^T - \frac{e_1 \mathcal{L}(\theta)}{\left \| T_2 \mathcal{L}(\theta) \right \|_2  } \leq 0.  
 \end{array} \end{equation}

\textbf{Preferred Pareto Optimality.}
To solve the constrained multi-objective subproblem with preference vector $u_i$ and $u_{i+1}$, we need to find an initial solution which is feasible or at least satisfies most constraints.
For a randomly generated solution $\theta$, we define the index set of all activated constraints as $\mathcal{K}_\epsilon(\theta) = \left\{k \mid \mathcal{Q}_k\left(\theta\right) \geq -\epsilon, k=0,\ldots, K-1\right\}$ and  $\mathcal{J}_\epsilon(\theta) = \left\{j \mid \mathcal{R}_j\left(\theta\right) \geq -\epsilon, j=0,\ldots, K-1\right\}$, where $\epsilon$ is a threshold.
We can find a valid descent direction $d_t$ to reduce the value of all activated constraints by solving:
\begin{equation}\label{eqNewP} 	
\begin{array}{c} \left(d_{t}, \alpha_{t}\right)=\arg \min _{d \in R^{M}, \alpha \in R} \alpha+\frac{1}{2}\|d\|^{2}_2, \\ 
\text { s.t. } \quad \nabla \mathcal{L}_{m}\left(\theta_{t}\right)^{T} d \leq \alpha, m=1, \ldots, M. \\
\nabla \mathcal{Q}_k(\theta_t)^{T} d \leq \alpha, k \in \mathcal{K}_\epsilon(\theta_{t}), \\ 
\nabla \mathcal{R}_j(\theta_t)^{T} d \leq \alpha, j \in \mathcal{J}_\epsilon(\theta_{t}).
\end{array} \end{equation}

The valid descent direction can be found by solving the constrained optimization problem in Eq. \ref{eqNewP}. 
However, the optimization problem itself is not scalable well for high dimensional decision space especially in deep neural networks. To solve the constrained optimization problem, we propose a scalable optimization method. 
We rewrite the optimization problem Eq. \ref{eqNewP} in its dual form. Based on the KKT conditions, we have the update direction as
\begin{equation}\label{dUp}
	\begin{array}{c} d_{t} \left( \omega_{m}, \beta_{k}, \gamma_{j} \right) =-\left(\sum_{m=1}^{M} \omega_{m}  \nabla \mathcal{L}_{m}\left(\theta_{t} \right) 
	\right. \\ \left. 	+\sum_{k \in \mathcal{K}_\epsilon(\theta_{t})}\beta_{k}  \nabla \mathcal{Q}_{k}\left(\theta_{t} \right)  
	 + \sum_{j \in \mathcal{J}_\epsilon(\theta_{t}) } \gamma_{j}\nabla \mathcal{R}_j(\theta_t ) \right),\end{array}
\end{equation}
where $\omega_{m} \geq 0 , \beta_{k} \geq 0 \text { and } \gamma_{j} \geq 0$ are the Lagrange multipliers for the linear inequality constraints. Therefore, the dual problem is:
\begin{equation}\label{dualP}
	\begin{array}{c}\max_{\omega_{m}, \beta_{k}, \gamma_{j}} -\frac{1}{2}\left\| d_{t} \left( \omega_{m}, \beta_{k}, \gamma_{j} \right)  \right\|^{2}_2 , \\  
	 
	\text { s.t. }  
	\sum_{m=1}^{M} \omega_{m} +\sum_{k \in \mathcal{K}_\epsilon(\theta_t)} \beta_{k}  + \sum_{j \in \mathcal{J}_\epsilon(\theta_t)} \gamma_{j} =1, \\ 

	 \forall m=1, \ldots, M, \forall  k \in \mathcal{K}_\epsilon(\theta_t), \forall j \in \mathcal{J}_\epsilon(\theta_t), \\ 
	 	\omega_{m} \geq 0,  \beta_{k} \geq 0, \gamma_{j} \geq 0.\end{array}
\end{equation}

As shown in Fig. \ref{figpre}, the preference vectors divide the objective space into different sub-regions,  and with the restricted of $u_0$, the space where auxiliary tasks hamper the performance of the main task will not be considered, the computation at training time decreases with a smaller exploration space.
The set of solutions for all subproblems would be in different sub-regions and represent different trade-offs among the tasks.

\begin{figure}
	
	\begin{subfigure}{0.46\linewidth}
		\includegraphics[width=\linewidth]{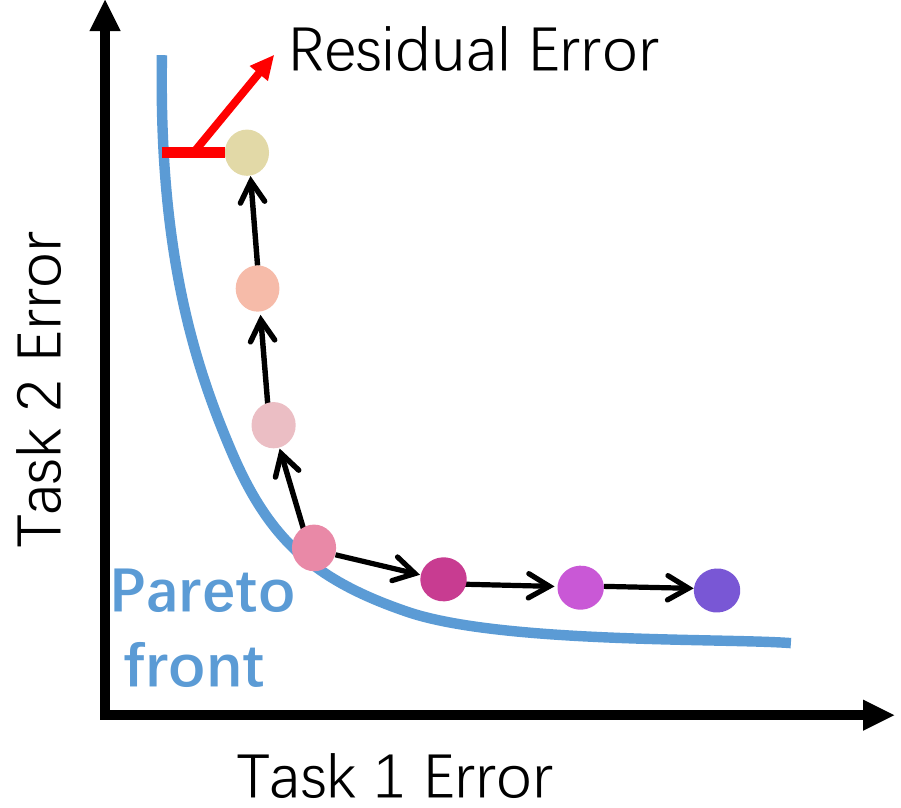}\label{ablation}
		\caption{Previous work}
	\end{subfigure}
	\begin{subfigure}{0.46\linewidth}
		\includegraphics[width=\linewidth]{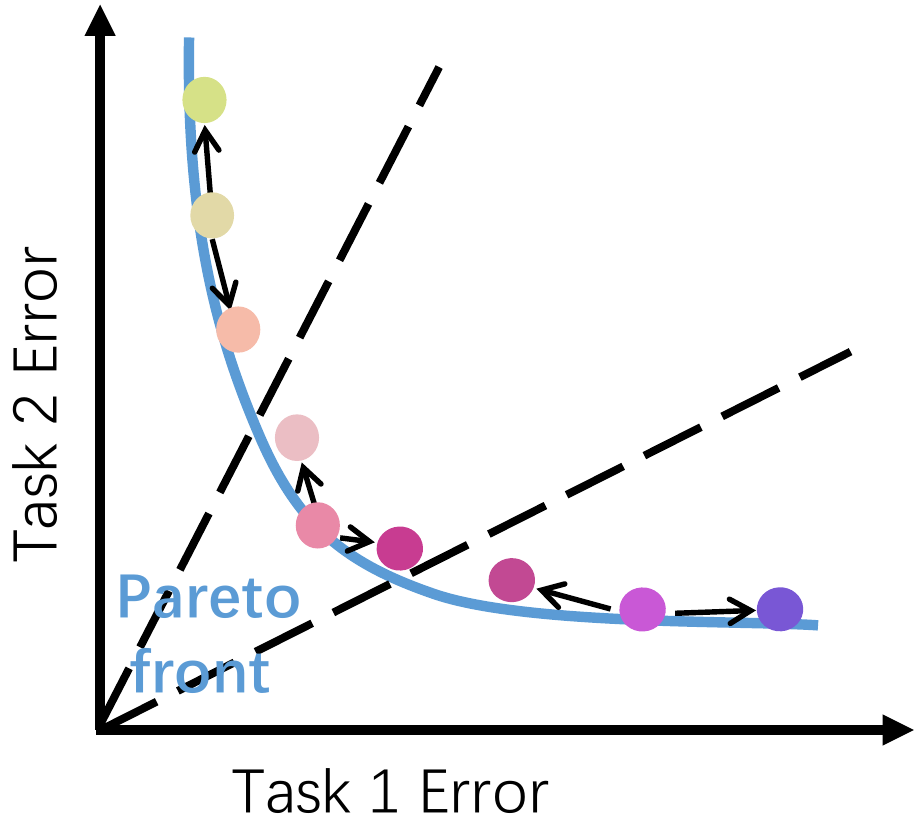}\label{entropy}
		\caption{PSST}
	\end{subfigure}
	\caption{Illustrative examples of residual error accumulation problem in previous Pareto exploration, and PSST alleviate this issue by preferred Pareto exploration. }\label{figproblem}
\end{figure}

\begin{figure}
	\begin{subfigure}{0.46\linewidth}
		\includegraphics[width=\linewidth]{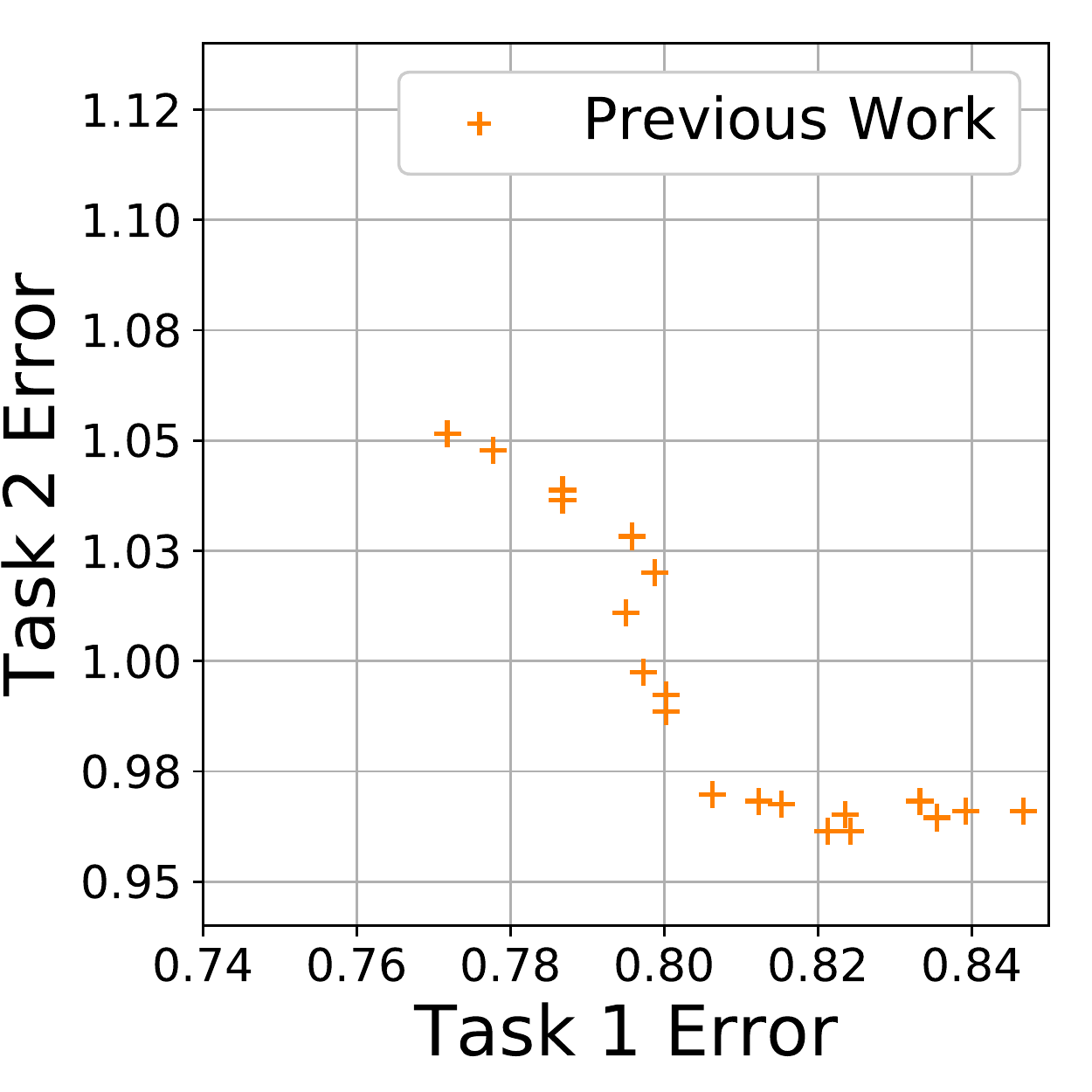}\label{ablation}
		\caption{}
	\end{subfigure}
	\begin{subfigure}{0.46\linewidth}
		\includegraphics[width=\linewidth]{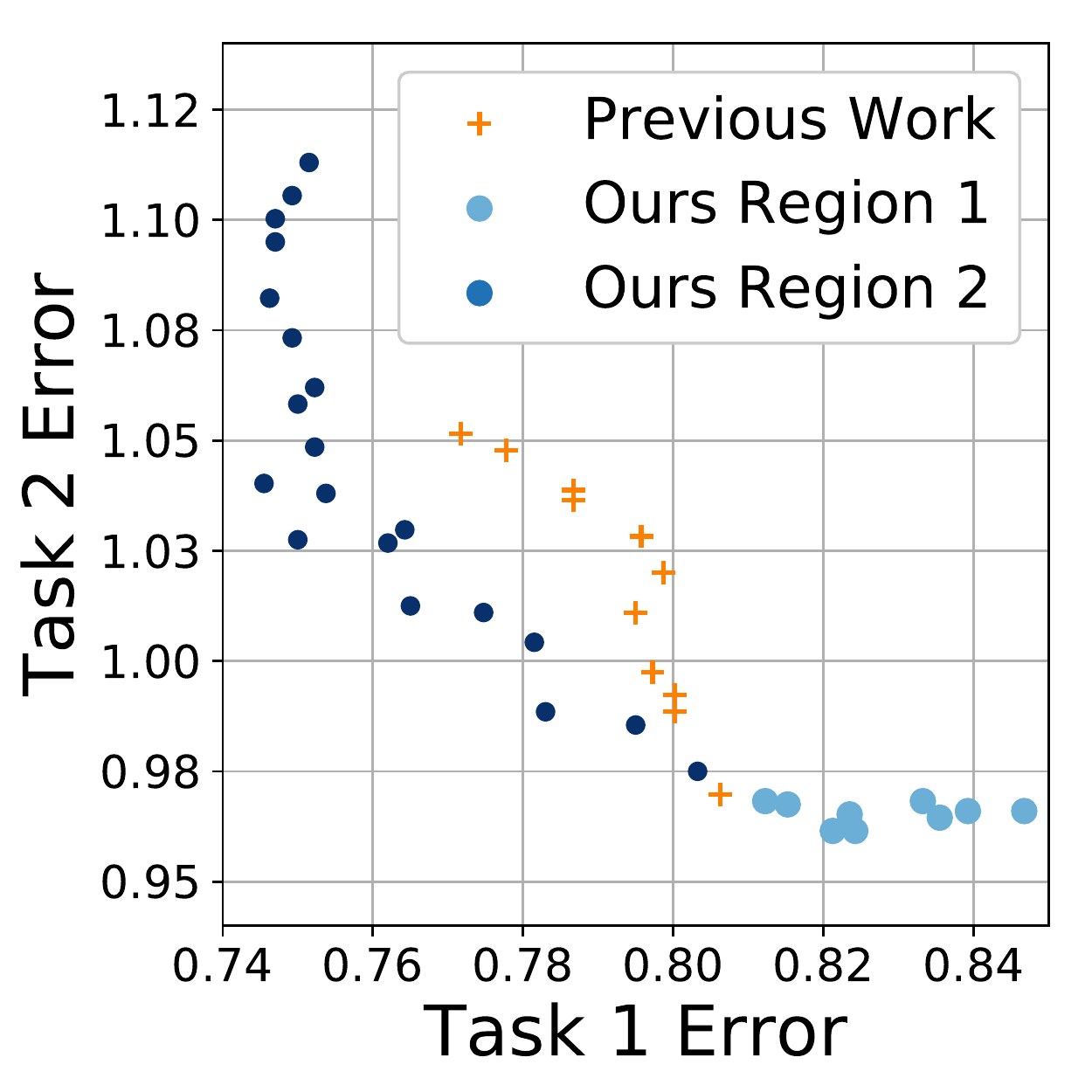}\label{entropy}
		\caption{}
	\end{subfigure}
	\caption{Residual error comparison between (a) previous work and (b) PSST, details in Section \ref{exper}.}\label{figExpproblem}
\end{figure}

\subsection{Preferred Pareto Exploration}\label{paretoEx}

Pareto exploration aims at finding several Pareto solutions.
Different from the grid search in the weighted sum of objectives, the Pareto exploration is more efficiency in finding Pareto solutions \cite{ma2020continuous}.
However, previous Pareto exploration has the residual error accumulation problem as shown in Figs. \ref{figproblem} and \ref{figExpproblem}. 
To avoid such residual error accumulation, we explore the sub-regions instead of the whole region.

\textbf{Find a Pareto solution.}
Firstly, 
our preferred Pareto exploration takes random parameters of neural network $\theta$ as input, and find a Pareto solution $\theta^*$ via direction $d_t$ in Eq.\ref{dUp}.

\textbf{Find gradient on the tangent plane at $\theta^*$. } 
Secondly, we explore the local Pareto set of Pareto solution $\theta^*$ 
by spawning new points $\theta_t$. 

\textbf{Lemma 3 \cite{hillermeier2001nonlinear}}: If $\theta^*$ is Pareto optimal, there will be a $\boldsymbol{\lambda} \in \mathbb{R}^{M}$ such that $\lambda_{m} \geq 0$, $\sum_{m=1}^{M} \lambda_{m}=1,$ and $\sum_{m=1}^{M} \lambda_{m} \nabla \mathcal L_{m}(\theta^*)={0}$. 

\textbf{Proposition \cite{hillermeier2001generalized}}: Assuming that $ \mathcal{L}\left(\theta^*\right)$
is smooth and $\theta^*$ is Pareto optimal, consider any smooth
curve ${\theta}_{(x)}:(-\epsilon, \epsilon) \rightarrow \mathbb{R}^{M}$ in the Pareto set and passing $\theta^*$ at $x = 0$, i.e., ${\theta}_{(0)}={\theta}^{*}$, then for $\exists \beta \in \mathbb{R}^{M}$ we have:
\begin{equation}
\begin{array}{cc}
\boldsymbol{H}\left(\theta^*\right) \frac{d\theta}{dx}(0)=\nabla \mathcal{L}\left(\theta^* \right)^{\top} \boldsymbol{\beta},
\end{array}
\end{equation}
where
$\boldsymbol{H}\left(\theta^*\right)=\sum_{m=1}^{M} \lambda_{m} \nabla^{2} \mathcal{L}_{m}\left(\theta^*\right)$,
and we use $\frac{d\theta}{dx}(0)$ as the update direction and calculate $\theta_{1} = \theta_{(0)} + \eta \frac{d\theta}{dx}(0)$. 
Solving such problem requires an efficient matrix solver. Similar to \cite{ma2020continuous}, we use Krylov subspace iteration methods.

\textbf{Explore a set of Pareto solutions (line 8 in Algorithm \ref{a1}).}
To explore a set of Pareto solutions based on $\theta^*$ in the region between $u_i$ and $u_{i+1}$, we initialize a queue $q \leftarrow\left[\theta^*\right]$ and a set $s\leftarrow\left[\theta^*\right]$.
Given a Pareto solution $\theta^*$, with gradient $\frac{d\theta}{dx}(0)$ on the tangent plane at $\theta^*$, we can have $\theta_{1}=\theta^*+\eta \frac{d\theta}{dx}(0)$.
Here, $\theta_{1}$ is the first-order approximation of a Pareto solution close to $\theta^*$, and we can have a Pareto solution $\theta_{1}^*$ via direction $d_t$ in Eq.\ref{dUp}. 
The queue and set collect all the Pareto solutions: $q \leftarrow\left[\theta_{t}^*\right]$ and $s\leftarrow\left[\theta_{t}^*\right]$. 
Once we cannot find any approximate point at $\theta^*$, we remove $\theta^*$ from queue $q$, and do the same to all points in the queue $q$.
Thus, we explore all the Pareto solutions in the region between $u_i$ and $u_{i+1}$. Pareto solutions in all regions are saved in set $s$.
The Pareto solution which has the best performance in the main task will be the final solution for a specific few-shot task.
We summarize the PSST in Algorithm \ref{a1}.

To summarize, the efficiency of our exploration algorithm comes from two sources: exploration on the tangent plane and termination with restricted preference region.  
The time cost of getting one tangent direction is $O(kn)$, $k$ is the iteration and $n$ is the size of input data, which scales linearly to the network size.

\begin{algorithm}[t]
	\caption{Few shot Learning via Pareto Self-supervised Training}
	\begin{algorithmic}[1]
			\State \textbf{Input:} A random initial neural network $\theta$.   
			\State Find a Pareto solution $\theta_{\pi0}^*$ via direction $d_t$ in Eq. \ref{eqP}
			\State Find $ \pi_0 = \arccos\frac{e_{1} \mathcal{L}(\theta_{\pi0}^*)}{\left \| T_2 \mathcal{L}(\theta_{\pi0}^*)\right \|_2}$, $u_0 = \left(\cos \pi_{0}, \sin \pi_{0} \right)$. 
			\State Calculate vectors $\left\{\boldsymbol{u}_{1}, \boldsymbol{u}_{2}, \ldots, \boldsymbol{u}_{K}\right\}$ via Eq. \ref{ui}.
			\ForAll{ $i=0 \text { to } K-1$}
		
			\State Randomly generate parameters $\theta^{(i)}$.
			
			\State Find Pareto solution $\theta^{(i)*}$ via direction $d_t$ in Eq. \ref{dUp}.

			\State $\boldsymbol{s}_{i} \leftarrow \text { PreferredParetoExploration }\left(\theta^{(i)*}\right)$

			\EndFor
			
			\State \textbf{Output:} The set of solutions for all subproblems $\left\{\boldsymbol{s}_{i} \mid i=1, \cdots, K\right\}$.
	\end{algorithmic}\label{a1} 
\end{algorithm}

\begin{table*}[]
	
	\centering 
	\begin{tabular}{l|c|c|c}
		\toprule[2pt]
		Models &	Backbone	&	1-shot	& 5-shot     \\ 
		\hline
		\hline
		MetaOptNet \cite{LeeMRS19}& ResNet-12 &62.64 $\pm$  0.61& 78.63 $\pm$  0.46 \\
		\quad  + rot, (\textit{SLA} \cite{lee20_sla}) & ResNet-12 &62.93 $\pm$  0.63  & 79.63 $\pm$  0.47    \\
		\quad  + rot\textbf{ + PSST} & ResNet-12 & \textbf{64.05$\pm$ 0.49  }  &  \textbf{80.24 $\pm$ 0.45   } \\
		
		\hline
		\hline
		
		CC   ~\cite{GidarisK18}      & WRN-28-10 &    61.09 $\pm$ 0.44  & 78.43 $\pm$ 0.33      \\
		\quad  + loc, (\textit{BF3S} \cite{GidarisBKPC19})  & WRN-28-10    & 60.71  $\pm$ 0.46  &  77.64 $\pm$ 0.34    \\
		\quad  + loc\textbf{ + PSST}  & WRN-28-10 &  \textbf{63.26 $\pm$ 0.44  } &\textbf{80.05 $\pm$ 0.34  }     \\
		\hline
		\quad  + rot, (\textit{BF3S} \cite{GidarisBKPC19})  & WRN-28-10 &  62.93 $\pm$ 0.45  &  79.87 $\pm$ 0.33   \\
		\quad  + rot\textbf{ + PSST} & WRN-28-10 &  \textbf{64.16 $\pm$ 0.44 } &\textbf{80.64 $\pm$ 0.32   }      \\
		\hline
		\hline
		ProtoNet \cite{SnellSZ17}& ResNet-18 &54.16 $\pm$  0.82&  73.68 $\pm$  0.65 \\
		\quad  + rot, (\textit{SSFSL} \cite{SuMH20}) & ResNet-18 &57.92 $\pm$  0.58  & 76.00 $\pm$  0.58   \\
		\quad  + rot\textbf{ + PSST} & ResNet-18 & \textbf{59.28 $\pm$ 0.46   }&   \textbf{77.18$\pm$ 0.43    }\\
		\hline
		\quad  + jig, (\textit{SSFSL} \cite{SuMH20}) & ResNet-18 &58.35 $\pm$  0.50   & 76.20 $\pm$  0.53   \\
		\quad  + jig\textbf{ + PSST} & ResNet-18 &\textbf{59.34 $\pm$ 0.45  }&   \textbf{77.27$\pm$ 0.43   } \\
		\hline 
		\quad  + jig + rot, (\textit{SSFSL} \cite{SuMH20}) & ResNet-18 &58.84 $\pm$  0.49    & 76.60 $\pm$  0.52    \\
		\quad  + jig + rot \textbf{ + PSST} & ResNet-18 &\textbf{59.52 $\pm$ 0.46  }&   \textbf{77.43$\pm$ 0.46  }  \\
		\bottomrule[2pt]
	\end{tabular} 
	\caption{ Average accuracy (\%) comparison with 95 confidence intervals before and after incorporating PSST into existing methods 
		on MiniImageNet. Best results are displayed in boldface.  
	}\label{PSST}
\end{table*}

\section{Experiments}\label{exper}
 
We design two experiments, which we call the sufficiency and necessity tests, to show the effectiveness and efficiency of our proposed PSST. 

In the necessity test, which focuses on demonstrate the effectiveness of our proposed PSST in few-shot auxiliary learning, we compare our approach with prior few-shot methods on the MiniImageNet and CIFAR-FS datasets respectively.  We call this experiment the necessity test as we use this experiment to establish that our proposed PSST, multi-objective optimization strategies rather than grid search 
in the weighted sum of objectives are indeed the source of efficiency in our method.

In the sufficiency test, we consider previous Pareto exploration method to show that our PSST is a fast and effective method.
We call this experiment the sufficiency test as it demonstrates our method is able to quickly explore Pareto sets and Pareto front in few-shot auxiliary learning.

\textbf{Datasets.}
We perform experiments on two few-shot datasets for necessity test: MiniImageNet \cite{VinyalsBLKW16} and CIFAR-FS \cite{BertinettoHTV19}. MultiMNIST \cite{sabour2017dynamic} is used for sufficiency test. 

\begin{table}[]	
	\centering 
	\resizebox{85mm}{35mm}{
		\begin{tabular}{c|c|c|c}
			\toprule[2pt]
			Models	&	Backbone	&	1-shot	& 5-shot     \\ \hline
			\hline
			MAML \cite{FinnAL17}&
			Conv-4-64&
			48.70 $\pm$ 1.84& 63.10 $\pm$ 0.92 \\
			Prototypical Nets \cite{SnellSZ17} &Conv-4-64 &
			49.42 $\pm$ 0.78 &68.20 $\pm$ 0.66 \\
			R2-D2 \cite{BertinettoHTV19} & Conv-4-64 & 48.70 $\pm$ 0.60 & 65.50 $\pm$ 0.60 \\
			LwoF \cite{GidarisK18} &
			Conv-4-64 &
			56.20 $\pm$ 0.86 & 72.81 $\pm$ 0.62 \\
			BF3S \cite{GidarisBKPC19}	&	Conv-4-64 & 54.83 $\pm$ 0.43 &71.86 $\pm$ 0.33   \\ 
			RelationNet \cite{Sung:RelationNet} &
			Conv-4-64 &
			50.40 $\pm$ 0.80 & 65.30 $\pm$ 0.70 \\
			GNN \cite{SatorrasE18} & Conv-4-64 & 50.30 & 66.40 \\ 
			TADAM \cite{oreshkin2018tadam} & ResNet-12 &
			58.50 $\pm$ 0.30& 76.70 $\pm$ 0.30 \\
			Munkhdalai et al. \cite{MunkhdalaiY17} &ResNet-12&
			57.10 $\pm$ 0.70 &70.04 $\pm$ 0.63 \\
			SNAIL \cite{Mishra:SNAIL}&
			ResNet-12&
			55.71 $\pm$ 0.99 &68.88 $\pm$ 0.92 \\
			Shot-Free \cite{RavichandranBS19} & ResNet-12 & 59.04 & 77.64 \\
			MetaOptNet \cite{LeeMRS19}& ResNet-12 &62.64 $\pm$  0.61& 78.63 $\pm$  0.46 \\
			Qiao et al. \cite{QiaoLSY18}	&	WRN-28-10&  61.76 $\pm$ 0.08 &77.59 $\pm$ 0.12   \\ 
			BF3S \cite{GidarisBKPC19}	&	WRN-28-10&  62.93 $\pm$ 0.45 &79.87 $\pm$ 0.33   \\ 
			Su et al. \cite{SuMH20}	&	WRN-28-10&  60.43 $\pm$ 0.58 &76.60 $\pm$ 0.29   \\ 
			AWGIM \cite{GuoC20} & WRN-28-10 & 63.12 $\pm$ 0.08 &78.40 $\pm$ 0.11 \\ 
			\hline
			{PSST}	&	Conv-4-64  & {57.04 $\pm$ 0.51} & {73.17 $\pm$ 0.48}   \\ 
			{PSST}	&	WRN-28-10&  \textbf{64.16 $\pm$ 0.44} &\textbf{80.64 $\pm$ 0.32}   \\ 
			
			\bottomrule[2pt]
		\end{tabular}
	} 
	\caption{Few-shot image classification average accuracy (\%) comparison with state-of-the-arts with 95\% confidence intervals on MiniImageNet.}\label{MiniCom}
\end{table}

\begin{table}[] 
	\centering
	\resizebox{85mm}{25mm}{
		\begin{tabular}{c|c|c|c}
			\toprule[2pt]
			Models	&	Backbone	&	1-shot	& 5-shot     \\ \hline
			\hline
			Prototypical Nets \cite{SnellSZ17}&
			Conv-4-64&
			62.82 $\pm$ 0.32 &79.59 $\pm$ 0.24 \\ 
			MAML \cite{FinnAL17} &
			Conv-4-64 &
			58.90 $\pm$ 1.90 &71.50 $\pm$ 1.00  \\ 
			RelationNet \cite{Sung:RelationNet} & Conv-4-64 &
			55.00 $\pm$ 1.00 & 69.30 $\pm$ 0.80   \\
			BF3S \cite{GidarisBKPC19} & Conv-4-64 &
			63.45 $\pm$ 0.31 & 79.79 $\pm$ 0.24   \\
			GNN \cite{SatorrasE18} &
			Conv-4-64 & 61.90  &
			75.30   \\ 
			R2-D2 \cite{BertinettoHTV19} &
			Conv-4-64 &
			60.00 $\pm$ 0.70 & 76.10 $\pm$ 0.60  \\ 
			Shot-Free \cite{RavichandranBS19} &ResNet-12& 69.20 $\pm$ 0.40& 84.70 $\pm$ 0.40  \\
			MetaOptNet \cite{LeeMRS19} &ResNet-12 &72.00 $\pm$ 0.70& 84.20 $\pm$ 0.50  \\
			BF3S \cite{GidarisBKPC19}	&	WRN-28-10&  76.09 $\pm$ 0.30 &87.83 $\pm$ 0.21   \\   
			\hline
			{PSST}& Conv-4-64 & {64.37 $\pm$ 0.33}  & {80.42 $\pm$ 0.32}  \\ 
			{PSST}& WRN-28-10 & \textbf{77.02 $\pm$ 0.38}  & \textbf{88.45 $\pm$ 0.35}  \\    
			\bottomrule[2pt] 
		\end{tabular}
	} 
	\caption{Few-shot image classification average accuracy (\%)
		comparison with state-of-the-arts with 95\% confidence intervals on CIFAR-FS.}\label{CIFARcom}
\end{table}

\textbf{Reduce the task conflict.} 
To verify the effectiveness of our proposed PSST, we embed it into three self-supervised tasks : location prediction (loc), rotation prediction (rot) and jigsaw puzzle (jig), and three widely used meta-learning baselines: MetaOptNet \cite{LeeMRS19}, Cosine Classifier (CC) \cite{GidarisK18} and ProtoNet \cite{SnellSZ17}. 
Table \ref{PSST} shows that for all cases, incorporating PSST leads to a significant improvement which demonstrates the effectiveness of our PSST. Specifically, the performance gain is 5.36\% on 1-shot and 3.75\% on 5-shot in ProtoNet with rotation prediction task and jigsaw puzzle task. 
Particularly,
the performance of BF3S method with location prediction task is lower than that without location prediction task in both 1-shot and 5-shot image classification task, which is caused by the \textit{task conflict}.
Incorporating PSST leads to a significant improvement .
We believe the reason is that location prediction task hampers the performance of BF3S, and our PSST effectively find a proper trade-off to reduce the task conflict between image classification and location prediction.

\textbf{Comparison with prior works.} In Tables \ref{MiniCom} and \ref{CIFARcom}, we compare our approach with prior few-shot methods on the MiniImageNet and CIFAR-FS datasets respectively. 
For our approach, we use CC and rotation prediction task, which gave the best results. In all cases we achieve state-of-the-art results surpassing prior methods. 

\textbf{Training time analysis.}  To further show the efficiency of PSST, we compare our approach with BF3S on the same GPU device as shown in Table \ref{t1}. For a single solution, the training time of two method is close.
For multiple solutions with different trade-offs, \eg 10 solutions, BF3S with different weights via grid search needs $249\times10^3$ iterations, and PSST only needs $60\times10^3$ iterations, which is much faster than BF3S due to the effective preferred Pareto exploration.

\begin{table}
	\begin{center}
		\centering
		\begin{tabular}{c|c|c|c|c}
			\toprule[2pt]
			Method & \multicolumn{2}{c|}{BF3S \cite{GidarisBKPC19}} & \multicolumn{2}{c}{PSST}        \\ \hline
			\hline
			Number of Solutions  & 1   & 10   & 1   & 10     \\ \hline
			Time (1 iteration)  & 1.02s   & 1.02s   & 1.05s   & 1.05s     \\ \hline
			Iterations ($\times10^3$)  & $24$   & $249$  & 24  & 60    \\ 
			\bottomrule[2pt]
		\end{tabular}
	\end{center} 
\captionof{table}{Training time on the same GPU device. Multiple solutions are based on different trade-offs.}\label{t1} 
\end{table}
 
In our sufficiency test, to evaluate our preferred Pareto exploration, similar to previous Pareto exploration work \cite{ma2020continuous}, we pick a subset of 2048 images from MultiMNIST.

\textbf{Effective region decomposition.}  
To demonstrate the effectiveness of our PSST, we assign PSST different initial parameters of neural network, and the results shows in Fig. \ref{figconvergence}.
All $u_0$ points are in the balance space where the overall performance of two tasks achieve good performance, however, which may not be the best performance for a specific task.
Given the $u_0$ points in the balance space, we can remove the space where the auxiliary tasks hamper the performance of the main task.
The smaller exploration region is one of the sources of efficiency in our method.
 
\textbf{Decrease residual error.} 
To analyze how PSST decreases residual error, we record the exploration trajectories of both previous method and our PSST as shown in Fig. \ref{figExpproblem}.
Compared with previous method, our trajectories are pushed towards its lower left, indicating a better approximated Pareto front.
We believe the reason is that the first-order approximation of a Pareto solution is closer to the Pareto front, and the 
exploration of $\theta^*$ is restricted in the sub-region with less residual error. 

\begin{figure}
	\begin{subfigure}{0.46\linewidth}
		\includegraphics[width=\linewidth]{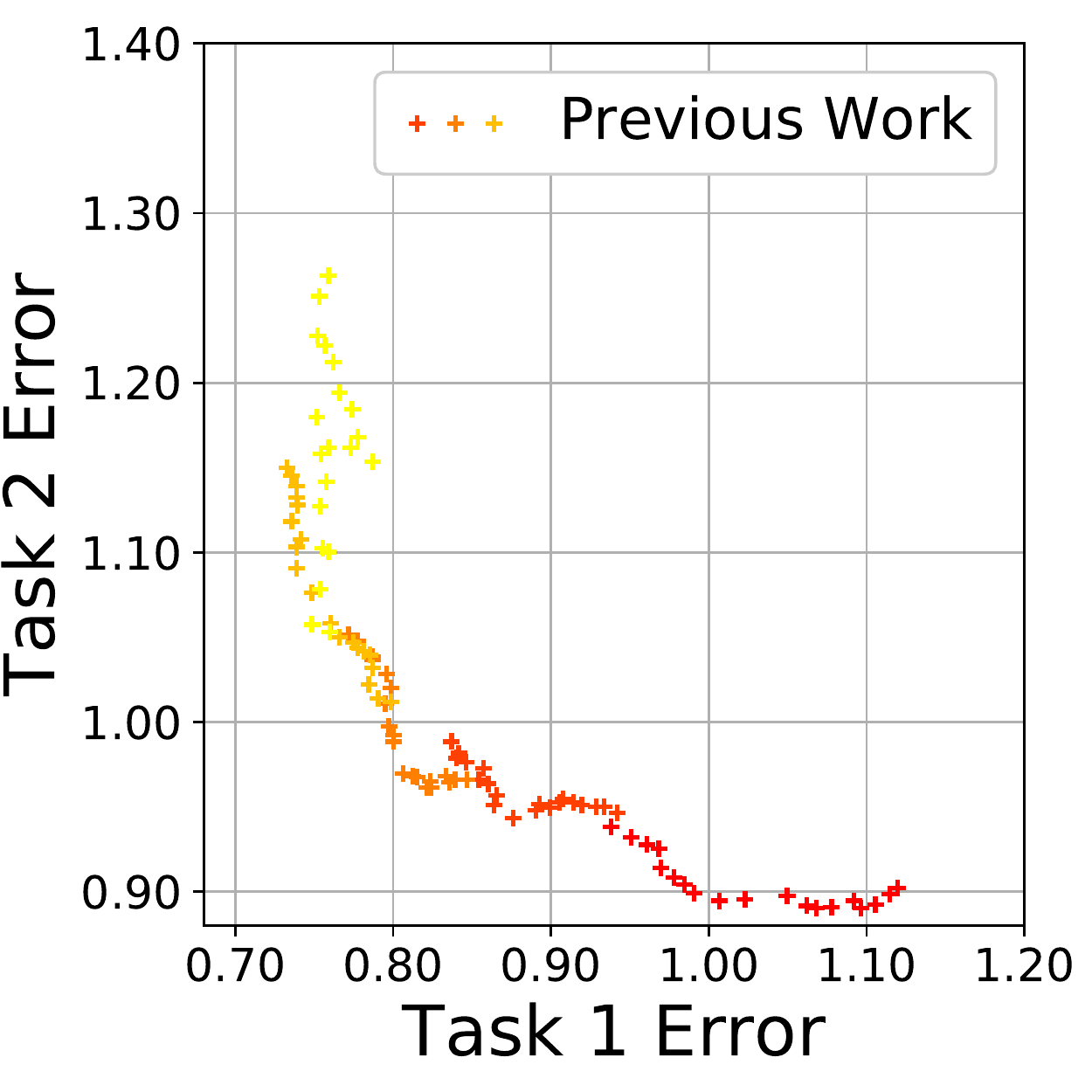}\label{ablation}
		\caption{}
	\end{subfigure}
	\begin{subfigure}{0.46\linewidth}
		\includegraphics[width=\linewidth]{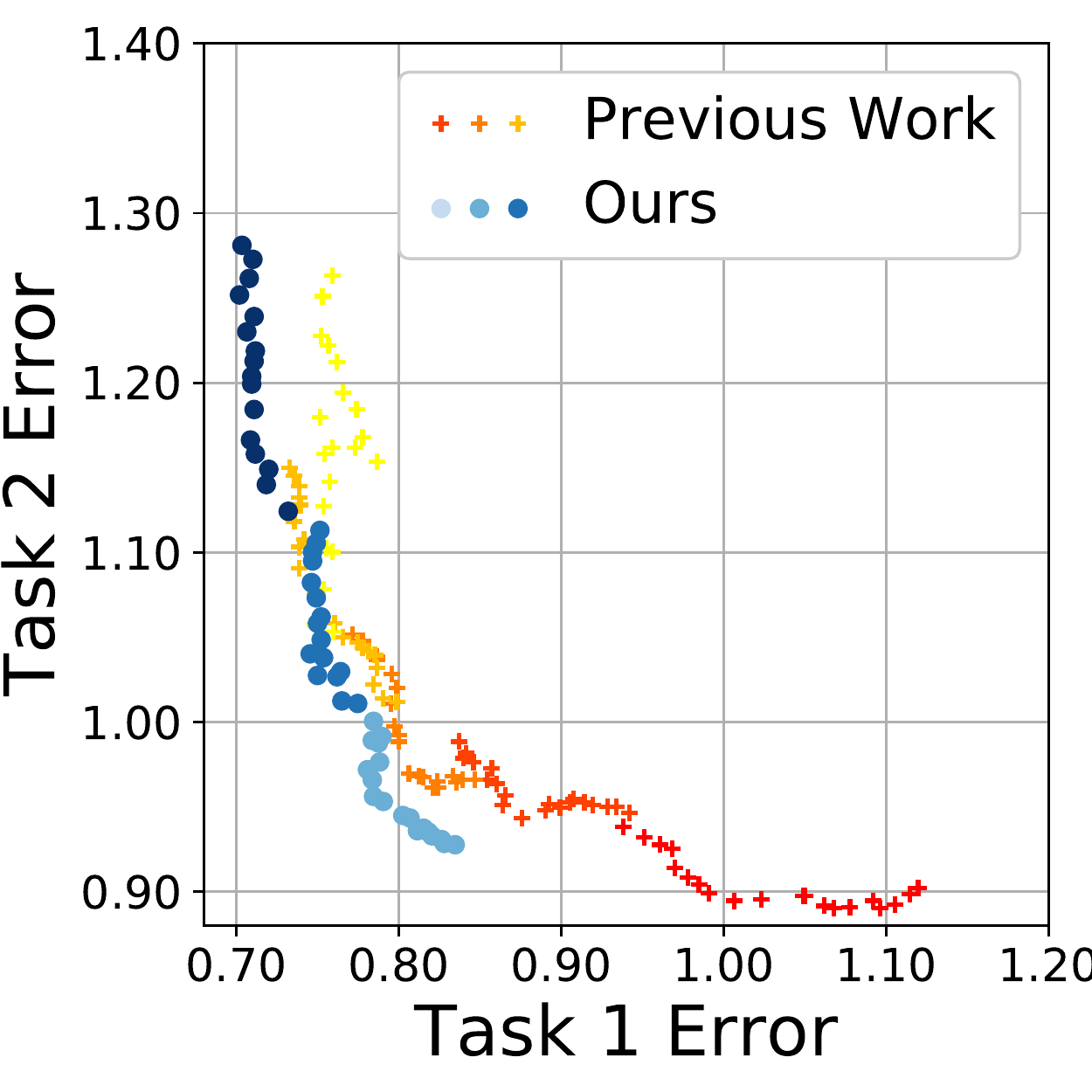}\label{entropy} 
		\caption{}
	\end{subfigure}
	\caption{Performance comparision between (a) previous Pareto exploration and (b) our PSST.}\label{figperform}
\end{figure}

\textbf{Performance comparison of Pareto exploration.}
As shown in Fig. \ref{figperform}, our method achieves a better performance on the main task with fewer explorations.
Our trajectories are pushed towards its lower left, indicating a better approximated Pareto front.
The training time of each iteration is the same for both methods, our PSST achieves better performance with 78 iterations, and previous work needs 130 iterations,
which demonstrates both effectiveness and efficiency of our PSST.  
The efficiency of PSST is from less exploration space and the effectiveness is from the decreasing residual error.

\section{Conclusion}
 
In this paper, we study the problem of task conflict in few-shot auxiliary learning. 
We propose a novel  
Pareto self-supervised training to reduce the task conflict in few-shot auxiliary learning.
We explicitly cast few-shot auxiliary learning as multi-objective optimization, with the overall objective of finding a Pareto optimal solution.
We decompose the few-shot auxililary problem into several constrained multi-objective subproblems with different trade-off preferences allowing better exploration of the frontier.  
Experiments demonstrate both efficiency and effectiveness of our proposed PSST: where the efficiency of PSST is from the less exploration space and the effectiveness is from the decreasing residual error.

\section{Acknowledgments}
The authors gratefully acknowledge funding support from the Westlake University and Bright Dream Joint Institute for Intelligent Robotics.

\end{document}